\documentclass[conference]{IEEEtran}
\IEEEoverridecommandlockouts
\usepackage{hyperref} 
\usepackage{cite}
\usepackage{amsmath,amssymb,amsfonts}
\usepackage{breqn}
\usepackage{algorithmic}
\usepackage{graphicx}
\usepackage{diagbox}
\usepackage{makecell}
\usepackage{textcomp}
\usepackage{xcolor}
\usepackage{subfigure}
\usepackage{subfloat}
\usepackage{booktabs}
\usepackage{color, colortbl}
\definecolor{GrayTop}{HTML}{ed7c31}
\definecolor{GrayMid}{HTML}{FFA492}
\definecolor{AdCream}{HTML}{FCFCD4}
\definecolor{AdOrange}{HTML}{ED7D31}
\definecolor{AdOrangeLight}{HTML}{FCE4D6}

\usepackage[noabbrev, capitalize]{cleveref}
\def\BibTeX{{\rm B\kern-.05em{\sc i\kern-.025em b}\kern-.08em
    T\kern-.1667em\lower.7ex\hbox{E}\kern-.125emX}}
    
\usepackage{etoolbox}
\makeatother
\makeatletter
\patchcmd{\@maketitle}
 {\addvspace{1\baselineskip}\egroup}
 {\addvspace{-1\baselineskip}\egroup}
 {}
 {}
\makeatother
\begin{document}

\title{Optimal activity and battery scheduling algorithm using load and solar generation forecasts
}

\author{
\IEEEauthorblockN{Yogesh Pipada Sunil Kumar}
\IEEEauthorblockA{
\textit{University of Adelaide}\\
Adelaide, Australia \\
\href{mailto:yogeshpipada.sunilkumar@adelaide.edu.au}{yogeshpipada.sunilkumar},}
\and
\IEEEauthorblockN{Rui Yuan}
\IEEEauthorblockA{
\textit{University of Adelaide}\\
Adelaide, Australia \\
\href{mailto:r.yuan@adelaide.edu.au}{r.yuan},}
\and
\IEEEauthorblockN{Nam Trong Dinh}
\IEEEauthorblockA{
\textit{University of Adelaide}\\
Adelaide, Australia \\
\href{mailto:trongnam.dinh@adelaide.edu.au}{trongnam.dinh},}
\and
\IEEEauthorblockN{S. Ali Pourmousavi}
\IEEEauthorblockA{
\textit{University of Adelaide}\\
Adelaide, Australia \\
\href{mailto:a.pourm@adelaide.edu.au}{a.pourm@adelaide.edu.au}}
}
\maketitle

\begin{abstract}
Energy usage optimal scheduling has attracted great attention in the power system community, where various methodologies have been proposed. However, in real-world applications, the optimal scheduling problems require reliable energy forecasting, which is scarcely discussed as a joint solution to the scheduling problem. The 5\textsuperscript{th} IEEE Computational Intelligence Society (IEEE-CIS) competition raised a practical problem of decreasing the electricity bill by scheduling building activities, where forecasting the solar energy generation and building consumption is a necessity. To solve this problem, we propose a technical sequence for tackling the solar PV and demand forecast and optimal scheduling problems, where solar generation prediction methods and an optimal university lectures scheduling algorithm are proposed.

\end{abstract}


\section{Introduction}
\subsection{Motivation}
\label{subsec:Motivation}
Even though infrequent, peak load is a cause for increased capital and operating expenses for power networks. The reason for this is twofold; the higher need for grid reinforcements and the use of more expensive fossil fuel-based generators to satisfy the peak loads (for a short duration). 
Thereby to manage the extra costs due to peak demand, network operators include a peak demand tariff (commensurate to the peak load) in the electricity bills of commercial customers. This motivates large-scale commercial customers such as universities and manufacturers to manage their demand better and invest in assets such as solar photovoltaic (PV) panels and stationary batteries that have the potential to shift their demand and reduce their electricity bills. A secondary effect of this is the reduction in CO\(_2\) emissions because of the lesser usage of fossil fuel generators, which has the potential to combat climate change and aid the process of decarbonization of our power networks.

However, optimal scheduling of (schedulable) loads and batteries (to minimize electricity costs) requires predictions of inflexible load (baseload), solar PV generation and the spot price of electricity. The first two variables majorly depend on weather conditions. Also, increasing renewable generation in the generation mix has led to increasing price volatility in the Australian national electricity market (NEM) which also creates a correlation between electricity prices and the weather \cite{downey2022untangling}. This makes designing such algorithms challenging as most commercial activities are planned over the mid to long term; hence require reliable predictions. Additionally, the scheduling problems are generally mixed-integer linear programs (MILPs) which are NP-hard and may be intractable depending on the formulation and problem size. 

Surrounding this premise, the IEEE Computational Intelligence Society (IEEE-CIS) partnered with Monash University (Victoria, Australia) to conduct a competition seeking technical solutions to manage Monash's micro-grid containing rooftop solar panels and stationary batteries~\cite{IEEECIS}. The main challenge was to develop an optimal scheduling algorithm for Monash's lecture theatres and operation of batteries to minimize their electricity bill, considering their baseload, solar generation and NEM electricity spot prices. To this end, the contestants were provided with actual time series data of the building loads without any lecture program (baseload) and solar generation from the micro-grid. So the contestants were expected to predict the baseload and solar generation by taking into account real-world weather data (also provided) for one month in the future. Following this, the contestants had to use actual electricity spot prices for the same duration along with these predictions for the optimal scheduling algorithm. 



\subsection{Related work} 
\label{subsec:Related Work}
We developed separate algorithms for the solar and baseload forecasts based on practical insights and the given data. Many approaches are available in the literature for both data types, a relatively mature and crowded research field. Therefore, the methods we have developed for this problem borrow themes from this mature literature. 

For solar predictions, the basic theme used is similar to the ``clear sky'' models for forecasting solar irradiance and thereby estimating PV generation \cite{Palani2017}. The general idea here is to create a baseline model for PV generation, assuming the sky is clear, meaning there is no cloud coverage or temperature variance. This baseline is then modified based on actual or expected weather conditions to estimate the actual PV generation. The literature around this idea is based on physical models of irradiance calculations, where equations are used to develop the baseline for a given geographical location \cite{Palani2017}. Newer methods use data-driven techniques such as time series forecasting and machine learning models~\cite{long2014analysis}. The data-driven methods are gaining popularity because of their robustness, speed and geographic adaptability. 

Intuitively, this method gives reliable solar forecasts because of solar generation's seasonal and diurnal nature. However, the main drawback is the lack of data specifically related to ``clear sky" days; Consequently, we use the most commonly occurring day's generation as baseline, which can be discovered by our previous work~\cite{yuan2021irmac}.

For the baseload forecasting, the theme used is the application of ensemble methods, as the current state of the art identifies these methods to be most accurate when compared to stand-alone methods \cite{raviv2015}. We use mainly a combination of random forest (RF), gradient boost (GB), autoregressive integrated moving average (ARIMA) and support vector machine (SVM), which are all well-studied and standard methods for time series forecasting \cite{HONG2016914}. Specifically, we apply different forecasting methods to different sub-series after disaggregating the load profiles using Seasonal and Trend decomposition using Loess (STL) due to their cyclic patterns \cite{Theodosiou2011forecasting}.

The optimal scheduling of distributed energy resources (DER) and controllable/uncontrollable loads for micro-grids comes under the class of problems commonly referred to as energy management problems. However, these classes of problems tend to be non-convex and non-linear in nature because of the constraints associated with the scheduled devices, e.g., scheduling uninterruptible loads such as lectures that cannot be stopped once started. As reviewed by the authors in~\cite{EMP_Review}, the majority of these problems are modeled via classical optimization approaches such as MILP~\cite{OptimalMILP} and mixed-integer non-linear programs (MINLPs)~\cite{MINLP_Eg}. Other approaches include dynamic programming~\cite{DP_Eg}, rule-based optimization and meta-heuristic algorithms~\cite{Meta_Eg}. 

Since these problems tend to be non-convex and non-linear, the major challenge associated with solving these problems is tractability and (in the case of meta-heuristic methods) convergence to a global optimum. The mathematical literature surrounding MILPs offers a variety of techniques to simplify these problems before solving them, and the availability of solvers such as Gurobi\textsuperscript{\textregistered} for MILPs make them a very attractive option for solving these problems. Also, unlike other models where the algorithms (may) converge to a local optimal solution, modern MILP solvers can obtain globally optimal solutions for this class of problems.


\subsection{Contributions}
\label{subsec:Contributions}
Based on the related work studied, this paper offers the following contributions to the body of knowledge: 
\begin{itemize}
    \item A solar forecasting algorithm using training data from refined motif (RM) discovery technique and use of an over-parameterized 1D-convoluted neural network (1D-CNN) implemented via residual networks (ResNet). To overcome the lack of ``clear sky'' data, we incorporated the work done previously by Rui Yuan et al.~\cite{yuan2021irmac} to identify RMs in the given solar generation dataset. An RM is the most repetitive pattern within a given time series, which can be extracted along with the exogenous variables (e.g., weather information) associated with this pattern. Using this as a baseline, we estimated solar generation by training an over-parameterized 1D-CNN. Some studies have shown that over-parameterization of CNNs can lead to better performance at the expense of longer training time \cite{Balaji2021, Bubeck2021,Power2021}. Therefore, a ResNet was implemented to develop a deeper NN but with faster computation time.
    \item An optimal micro-grid scheduling algorithm is solved based on real-world data for a university-based application. To the best of our knowledge, there has not yet been a study to co-optimize lectures schedule (and associated resources) and battery operation (with PV panels). Therefore, in this paper, we have developed and tested our algorithm using real-world data and practical case instances provided by Monash University. Given the large problem size (one-month schedule) and the presence of a quadratic term in the objective function for the cost of peak demand, we proposed a two sub-problem approach to formulate a tractable problem. The first sub-problem was used to limit the peak demand throughout the month, eliminating the quadratic term from the objective. Then, the peak demand was used to solve the second sub-problem to minimize the total electricity costs.
\end{itemize}

The rest of the paper is structured as follows. Section \ref{sec:Background} introduces the data, information and problem requirements. Section \ref{sec:Methodology} proposes the forecasting and optimization methodologies. Section \ref{sec:Results and Discussion} presents the numerical results of the scheduling algorithm based on time series forecasting. Finally, we conclude the paper section \ref{sec:Conclusion}.


\section{Background Information}
\label{sec:Background}
This section describes the competition requirements and data provided by the organizers.


\subsection{Problem statement}
\label{subsec:Problem Statement}
This section describes the scheduling constraints and objective function provided by the competition organizers and has been adapted from~\cite{IEEECIS}. We were required to develop prediction algorithms for the baseload of six buildings in Monash University and the solar generation of PV panels connected to them. After this, we had to use these predictions to optimally schedule lecture activities and battery operation while minimizing the electricity costs (i.e., electrical energy consumption cost plus peak demand charge). We were allowed to consider the electricity prices as known parameters to simplify the problem and focus on predicting solar generation and baseload.  

For each lecture activity, we are provided with several small or large rooms needed, the electrical power consumed per room and the duration of the activities (in steps of 15 minutes). We are also provided with a list of precedence activities, i.e., activities that must be performed at least one day before the activity in question. For the batteries, we are provided with the maximum energy rating, i.e., state of charge (SOC), the peak charge/discharge power and charge/discharge efficiency. 

The scheduling must start from the first Monday of the month, and each activity must; 1) take place within office hours (9:00--17:00), 2) happen once every week and recur on the same day and time every week. Also, the number of rooms (associated with each activity) allocated at a given time interval must not exceed the total number of rooms in the six buildings. The batteries 
can operate in one of three states; charging, discharging or idle, which is determined by the scheduling algorithm. However, the battery power must be at the peak charge/discharge capacity once operated.

\subsection{Dataset information}
\label{subsec:Data set}
The organizers provided us with datasets for the electricity consumption of the six buildings and their connected solar panels~\cite{IEEECIS}. About two years' worth of data was provided with a granularity of 15 minutes. However, the datasets contained many missing points, especially the demand profiles with consecutive weeks of unrecorded data, as shown in \cite{gitlabRay}.
Information regarding COVID-19 lockdown and how it influenced the forecast and earlier timelines were also provided to aid in incorporating the effect of COVID-19 pandemic on baseload in our models. ERA5 climate data was also provided as exogenous variables for the time series forecasting via a partnership with OikoLabs~\cite{IEEECIS}. 
Also, we were allowed to use electricity price information from the Australian NEM~\cite{NEM_Data} as a parameter for objective function calculation. The input parameters used for solar and building forecasting are summarized in Table~\ref{tab: inputs-sensitivity}.
\begin{table}[!h]
    \centering
    \caption{Input data used for forecasting}
    \label{tab: inputs-sensitivity}
    \resizebox{0.5\textwidth}{!}{
    \begin{tabular}{|l|*{2}{c|}}
        \hline
        \rowcolor{AdOrange} \backslashbox{Weather inputs (15mins)}{Forecast outputs (15mins)}
        &\makecell{Building \\(15mins)}&\makecell{Solar\\ (15mins)}\\\hline
        \rowcolor{AdOrangeLight}Temperature ($^\circ$C) & \checkmark & \checkmark \\\hline
        \rowcolor{AdCream}Dewpoint temperature ($^\circ$C) &\checkmark &\\\hline
        \rowcolor{AdOrangeLight}Wind speed (m/s) & &\checkmark \\\hline
        \rowcolor{AdCream}Relative humidity (0-1) & \checkmark & \checkmark\\\hline
        \rowcolor{AdOrangeLight}Surface solar radiation (W/$m^2$) &\checkmark &\checkmark\\\hline
        \rowcolor{AdCream}Total cloud cover (0-1) & &\checkmark\\\hline
        \rowcolor{AdOrangeLight}Occupancy  (0-1) & \checkmark&\\\hline
        \rowcolor{AdCream}Annual harmonics (0-1) & &\checkmark\\\hline
    \end{tabular}
    }
\end{table}
\subsection{Instance information}
\label{subsec:Instance}

To test the algorithms each team developed, the competition organisers provided ten problem instances. Each instance provided information regarding the number of activities to be scheduled and the specifics of each activity; duration of the activity in time steps, power consumed per room for each time interval, number of large/small rooms required and a precedence list. The precedence list contains the activities that must happen one day before the specified task. Depending on the number of activities in each instance, they were divided into two sub-categories; large (200 activities) and small (50 activities) instances. We report simulation results for one large and one small instances in section~\ref{sec:Results and Discussion}.


\section{Methodology}
\label{sec:Methodology}
This section describes the methodology developed to solve the problem described in~\ref{subsec:Problem Statement}. Mainly, we outline our data cleaning strategies, forecasting methodology and scheduling algorithms used in the competition.


\subsection{Solar generation forecast}
\label{subsec:Solar generation forecast}

\begin{figure}[!ht]
\centerline{\includegraphics[clip, trim=3.3cm 4.3cm 2.75cm 3.95cm,width=0.95\linewidth]{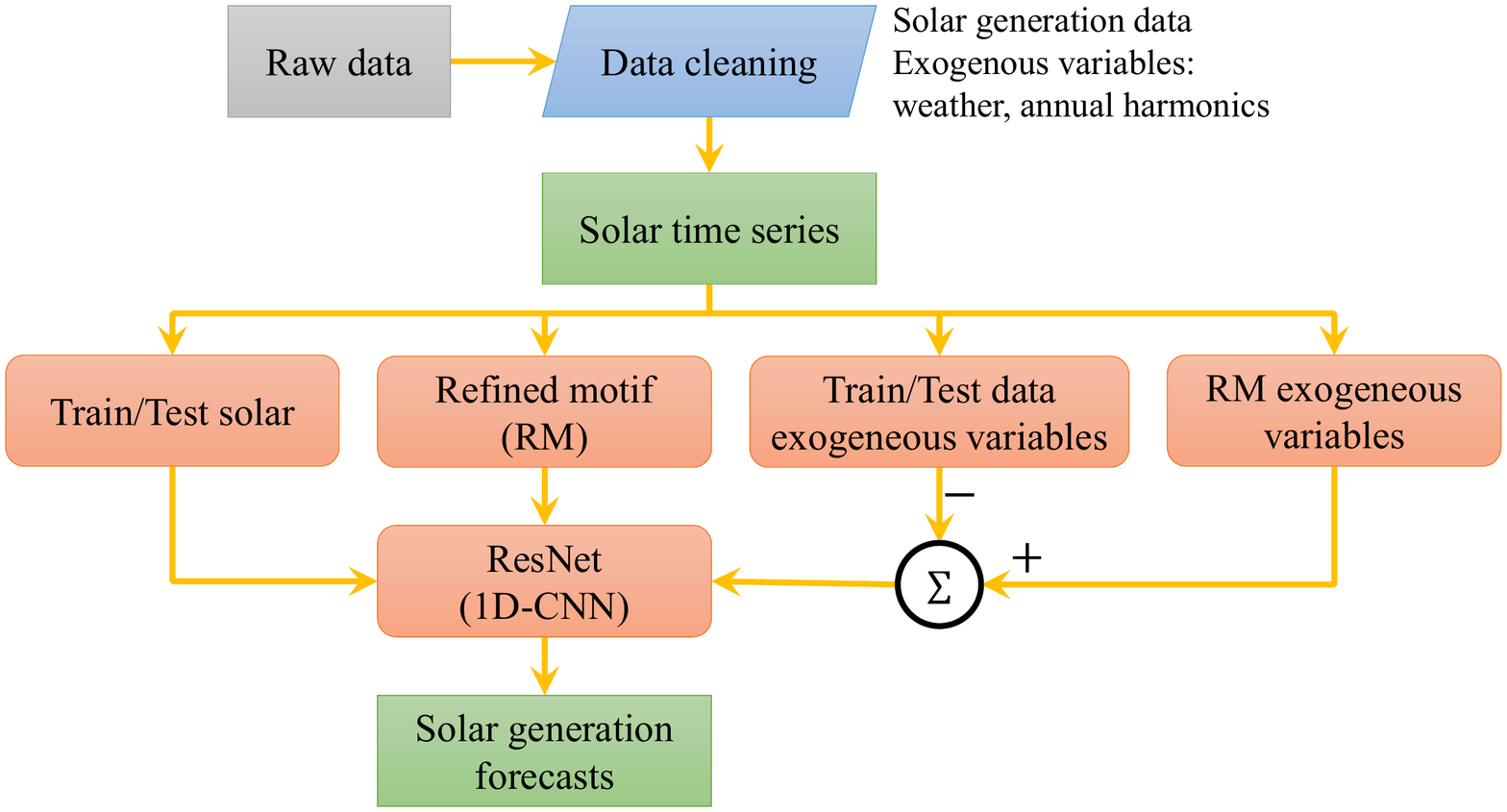}}
\caption{Block diagram of the proposed solar generation prediction model}
\label{fig:solar}
\end{figure}

The solar generation forecast methodology is depicted in Fig.~\ref{fig:solar}. The solar data was mainly clean, with no outliers and a few missing points, which were removed during the cleaning process. Since solar time series data is repetitive in nature, we used the method described in~\cite{yuan2021irmac} to extract the RM (most repetitive patterns) and its associated weather conditions. The intuitive idea is to train a neural network (NN) to reshape this RM with respect to the difference in weather conditions between the RM weather conditions and the actual data weather conditions. 

\subsection{Baseload forecast}
\label{subsec:baseload forecast}

\begin{figure}[!ht]
\centerline{\includegraphics[clip, trim=4.5cm 4.2cm 0.4cm 1.5cm,width=0.95\linewidth]{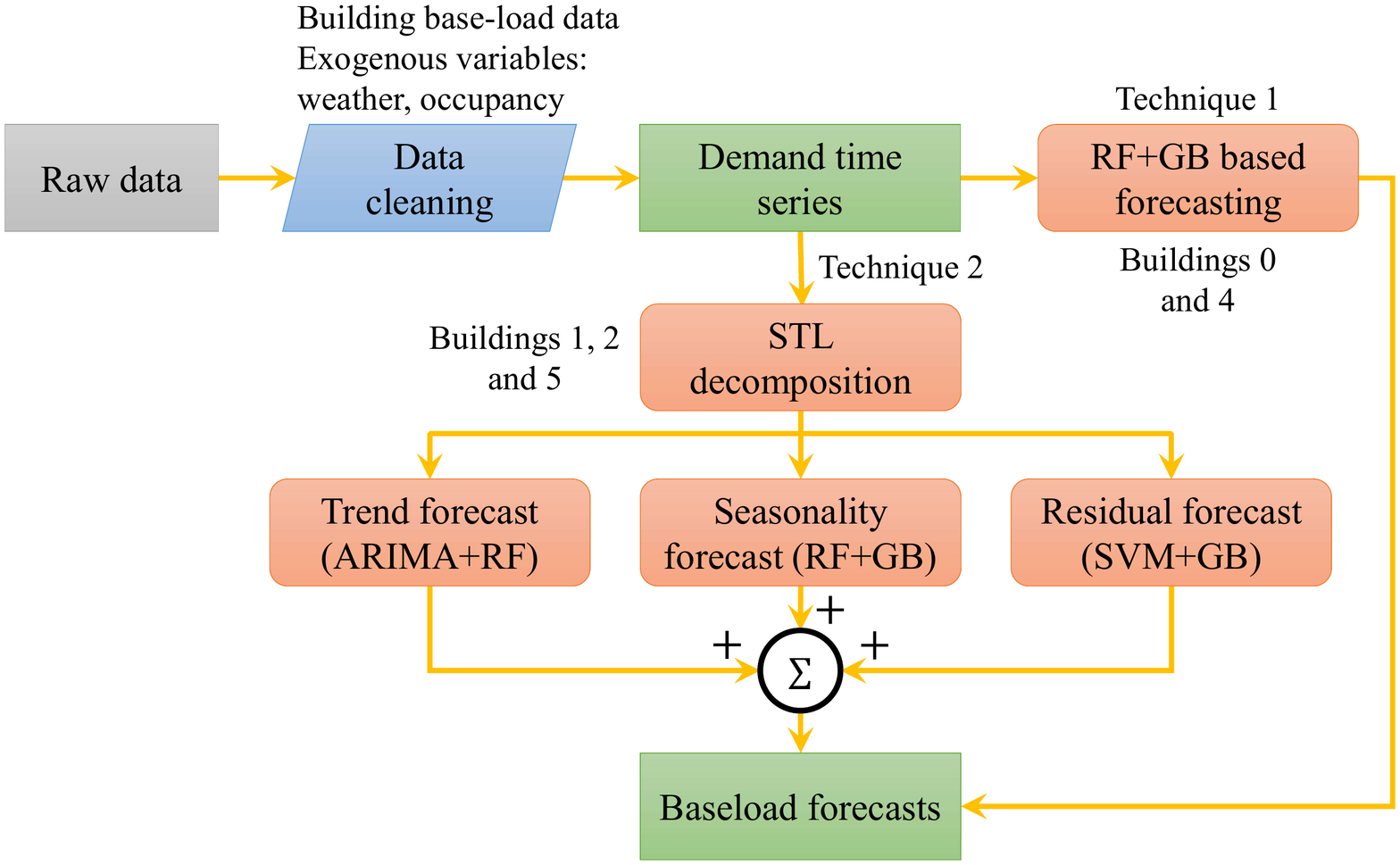}}
\caption{Block diagram of the proposed baseload prediction model}
\label{fig:base_load}
\end{figure}

The block diagram of the methodology used for baseload forecasting is shown in Fig.~\ref{fig:base_load}, which demonstrates the use of different ensemble forecasting methods for different buildings. We applied the same procedure for the data cleaning process as in the PV generation forecast. However, since there were more missing points in the demand profiles, which lasted for many weeks consecutively, we used the average value from the previous week for the long-period missing data. The choice of ensemble methods was achieved through the use of the Python sklearn package's voting regressor function \cite{SklearnVoting}. For this dataset, this resulted in a combination of RF, SVM, ARIMA and GB methods for best accuracy. In terms of the length of training data, due to the highly dynamic nature of demand profiles, which was exacerbated during the COVID-19 period, we only used 2 months of historical data for training.

Due to the seasonal/cyclic patterns demonstrated in building 1, 2 and 5, as shown in \cite{gitlabRay}, we utilized STL decomposition to capture these properties by disaggregating the load profiles into three components, namely trend, seasonality and residual. We then trained each profile with different methods depending on the performance of the historical data. In the end, the predictions from these three sub-series were combined to obtain the final building demand forecast. For buildings 0 and 4 with no clear cyclic patterns, we observed that either RF or GB (without STL decomposition) were sufficient to capture their repetitive nature. Lastly, for building 3, historical data showed that the load only took on a finite set of discrete values; thus, we fixed the predictions to the median.


\subsection{Optimal scheduling algorithm}
\label{subsec:Optimal scheduling}

\begin{figure}[!ht]
\centerline{\includegraphics[clip, trim=3.6cm 9.3cm 1.2cm 3.5cm,width=0.95\linewidth]{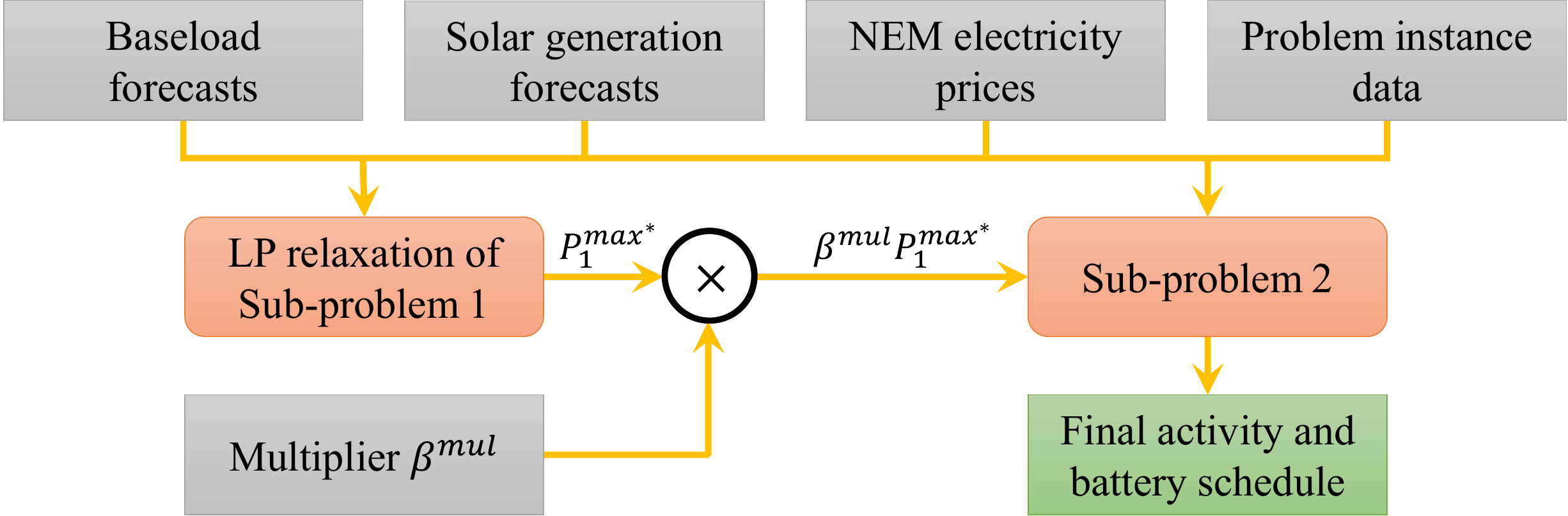}}
\caption{Block diagram of the proposed optimal scheduling algorithm}
\label{fig:scheduling_alg}
\end{figure}

We developed a MIP-based optimal scheduling problem to represent all the scheduling constraints, where the objective function was provided by the organizers as follows: 
\begin{flalign}
\label{eq:QP objective}
O(P_{t}) = \frac{0.25}{1000}\sum_{t \in \mathcal{T}}\, P_{t} \, \lambda^{\text{aemo}}_{t} + 0.005 \, \left(\max_{t \in \mathcal{T}} P_{t}\right)^2 &
\end{flalign}

Here \(P_{t}\) and \(\lambda^{\text{aemo}}_{t}\) are the net demand and AEMO electricity spot price at time \(t\) over the time horizon \(\mathcal{T}\) respectively. Therefore, it can be seen that the problem is a MIQP problem, which is NP-hard and hence most likely intractable given the size of this problem (\(\|\mathcal{T}\| = 2880\)). 
Hence, to improve the tractability, we reformulated the MIQP problem into two MILP sub-problems, as demonstrated in Fig.~\ref{fig:scheduling_alg}. The first sub-problem is used to estimate an upper bound for the maximum demand (\(P^{\text{max}}_1 = \max_{t \in \mathcal{T}} P_{t}\)), which is used to remove the quadratic term from the objective function in the second sub-problem. 

The scheduling is developed for a time duration \(\Delta t\) with \(P^{\text{base}}_t \text{ and } P^{\text{solar}}_t\) being the baseload and solar generation forecasts, respectively. For a set of batteries \(b \in \mathcal{B}\) over the time horizon \(t \in \mathcal{T}\), we define \(E^{\text{bat}}_{b, t}\) as battery SOC in kWh, \(P^{\text{bat}}_{b, t}\) is the battery power in kW, \(E^{\text{cap}}_b\) is the battery capacity in kWh, \(B^{\text{c}}_{b}/B^{\text{d}}_{b}\) is the battery charge and discharge power in kW, \(u^{\text{c}}_{b, t}/u^{\text{d}}_{b, t}\) are the battery charge/discharge status, and \(\eta^{\text{c}}_{b,t}/\eta^{\text{d}}_{b,t}\) are the charging/discharging efficiencies. For a set of activities \(a \in \mathcal{A}\) over the same time horizon, we define \(A^{\text{kW}}_a\) as the power required per room for the activity in kW, \(R^{\text{large}}_{a}/R^{\text{small}}_{a}\) are the number of large/small rooms needed for the activity, \(P^{\text{sched}}_{t}\) be the total power scheduled in kW, and \(u^{\text{start}}_{a,t}/u^{\text{active}}_{a,t}\) are binary variables which track the activity start time and active time, respectively. \(\mathcal{D}_{a}\) is the duration set for the activity, and \(\mathcal{P}_{a}\) is the set of activities that must precede activity \(a\). Additionally, we define \(\mathcal{T}_{n}\) as the set of time intervals where activities must not be active, \(\mathcal{T}_f\) is the set of time intervals for the first week and \(L^{\text{day}}_{t}\) corresponds to the day of the time interval \(t\). We present the equations used for the two sub-problems in the below sections.

\subsubsection{Sub-problem 1}
\label{subsubsec:Sub-problem 1}
\begin{subequations}
\begin{flalign}
\label{eqref:Objective 1}
\min_{\psi_1} P^{\text{max}}_{1} &
\end{flalign}
\begin{flalign}
&\textbf{Subject to,} \nonumber&
\end{flalign}
\begin{flalign}
\label{eqref:Max power constraint}
0 \leq P_{t}\leq P^{\text{max}}_{1} \quad \forall \, t \in \mathcal{T}& 
\end{flalign}
\begin{flalign}
\label{eqref:SOC limit}
0 \leq E^{\text{bat}}_{b, t} \leq E^{\text{cap}}_{b} \quad \forall \, t \in \mathcal{T}, \, b \in \mathcal{B}&
\end{flalign}
\begin{flalign}
\label{eqref:Charge-Discharge}
u^{\text{c}}_{b, t} + u^{\text{d}}_{b, t} \leq 1 \quad \forall \, t \in \mathcal{T}, \, b \in \mathcal{B}&
\end{flalign}
\begin{flalign}
\label{eqref:Battery power}
P^{\text{bat}}_{b, t} = u^{\text{c}}_{b, t}\,\frac{B^{\text{c}}_{b}}{\sqrt{\eta_{b}^{\text{c}}}} - u^{\text{d}}_{b, t}\,B^{\text{d}}_{b}\,\sqrt{\eta_{b}^{\text{d}}} \quad \forall \, t \in \mathcal{T}, \, b \in \mathcal{B} &
\end{flalign}
\begin{flalign}
\label{eqref:Battery capacity}
E^{\text{bat}}_{b, -1} = E^{\text{cap}}_{b} \quad \forall \,b \in \mathcal{B}&
\end{flalign}
\begin{flalign}
\label{eqref:SOC value}
E^{\text{bat}}_{b, t} = E^{\text{bat}}_{b, t-1} + \Bigg(u^{\text{c}}_{b, t}\, B^{\text{c}}_{b} - u^{\text{d}}_{b, t}\, B^{\text{d}}_{b} \Bigg) \Delta t \quad \forall \, t \in \mathcal{T}, \, b \in \mathcal{B}& 
\end{flalign}
\begin{flalign}
\label{eqref:Scheduled power}
P^{\text{sched}}_{t} = \sum_{a \in \mathcal{A}} u^{\text{active}}_{a,t} \, A^{kW}_a\, \Bigg(R^{\text{small}}_{a} + R^{\text{large}}_{a}\Bigg) \quad \forall \, t \in \mathcal{T}&
\end{flalign}
\begin{flalign}
\label{eqref:Net export}
P_{t} = P^{\text{base}}_{t} - P^{\text{solar}}_t + P^{\text{sched}}_{t}+ \sum_{b \in \mathcal{B}}P^{\text{bat}}_b \quad \forall \, t \in \mathcal{T} &
\end{flalign}
\begin{flalign}
\label{eqref:Room requirement}
\sum_{a \in \mathcal{A}} u^{\text{active}}_{a, t}\, R^{\text{large}}_{a} \leq H^{\text{large}}; \sum_{a \in \mathcal{A}} u^{\text{active}}_{a, t}\, R^{\text{small}}_{a} \leq H^{\text{small}} \; \forall t \in \mathcal{T}&
\end{flalign}
\begin{flalign}
\label{eqref:Non active time}
u^{\text{active}}_{a, t} = 0 ; u^{\text{start}}_{a, t} = 0 \quad \forall \, t \in \mathcal{T}_{n}, \, a \in \mathcal{A}& 
\end{flalign}
\begin{flalign}
\label{eqref:Happen once a week}
\sum_{t \in \mathcal{T}_f} u^{\text{start}}_{a, t} = 1 \quad \forall \, a \in \mathcal{A} &
\end{flalign}
\begin{flalign}
\label{eqref:Duration constraint A}
u^{\text{start}}_{a, t} \leq u^{\text{active}}_{a, t + k} \: \forall \, & t \in \{0, 1,\dots, \left|\mathcal{T}\right| - \left|\mathcal{D}_{a}\right| - 1\}, \, k \in \mathcal{D}_{a}, a \in \mathcal{A}&
\end{flalign}
\begin{flalign}
\label{eqref:Duration constraint B}
\sum_{t \in \mathcal{T}_f} u^{\text{active}}_{a, t} = \left| \mathcal{D}_{a} \right| \quad \forall \, a \in \mathcal{A}&
\end{flalign}
\begin{flalign}
\label{eqref:Precedence constraints}
\sum_{t \in \mathcal{T}_f} \left(u^{\text{active}}_{a, t}\, L^{\text{day}}_{t} - u^{\text{active}}_{k, t}\, L^{\text{day}}_{t}\right) \geq 1 \; \; \forall \, k \in \mathcal{P}_{a}, a \in \mathcal{A}&
\end{flalign}
\begin{flalign}
\label{eqref:Start at same time every week}
u^{\text{start}}_{a, t} = u^{\text{start}}_{a, t - \alpha} \quad \forall \, t \in \mathcal{T}_{f}^{c}, \, a \in \mathcal{A} &
\end{flalign}
\end{subequations}

\noindent where \(\psi_1 = \{P, P^{\text{bat}}, P_1^{\text{max}}, E^{\text{bat}},u^{\text{c}}, u^{\text{d}}, u^{\text{active}}, u^{\text{start}} \}\). Equation~\eqref{eqref:Max power constraint} helps find the peak demand of the schedule and prevents energy export to the grid. Equations~\eqref{eqref:SOC limit}--\eqref{eqref:SOC value} represent battery related constraints. Equations~\eqref{eqref:Scheduled power}--\eqref{eqref:Net export} are used to calculate the net energy imported from the grid, and equation~\eqref{eqref:Room requirement} ensures the number of rooms engaged through activities is within the limit. Equation~\eqref{eqref:Non active time} ensures activities do not start outside office hours and before the first Monday of the month, equation~\eqref{eqref:Happen once a week} ensures activities happen once a week, equations~\eqref{eqref:Duration constraint A}--\eqref{eqref:Duration constraint B} ensure activities are active for the necessary duration after starting, equation~\eqref{eqref:Precedence constraints} satisfies that all precedence (prerequisite) activities happen a day before and equation~\eqref{eqref:Start at same time every week} ensures activities recur at the same time every week. The objective here is to minimize peak demand while satisfying all the scheduling constraints.  

Since this problem is only used to solve for the upper bound maximum demand \(P^{\text{max}}_1\), it is not necessary to solve the full problem. We use the property that the solution of the LP relaxation of a minimization-based MILP problem is its lower bound (\(y^{*}_{\text{MILP}} \geq y^{*}_{\text{LP}}\))~\cite{Vielma2015} to just solve the LP relaxation and use this for sub-problem 2, thereby improving tractability of our method.


\subsubsection{Sub-problem 2}
\label{subsubsec:Sub-problem 2}
\begin{subequations}
\begin{flalign}
\min_{\psi_2} \frac{0.25}{1000}\sum_{t \in \mathcal{T}}\, P_{t} \, \lambda^{\text{aemo}}_{t}  &
\end{flalign}
\begin{flalign}
&\textbf{Subject to, Equations~}\eqref{eqref:SOC limit} \text{\---} \eqref{eqref:Precedence constraints}\nonumber&
\end{flalign}
\begin{flalign}
\label{eqref:Load limit}
0 \leq P_{t} \leq \beta^{\text{mul}} \,P^{\text{max}^{*}}_{1} &
\end{flalign}
\end{subequations}

\noindent where \(\psi_2 = \left(\psi_1 \setminus \{P^{\text{max}}_{1}\} \right)\). Using equation~\eqref{eqref:Load limit}, we ensure that peak demand does not exceed \(\beta^{\text{mul}} P^{\text{max}^{*}}_1\) limiting the demand charge. Intuitively, \(\beta^{\text{mul}} \geq 1\) is a multiplier used to change the tightness of the MILP, creating a trade-off between computation time and electricity cost; higher values of \(\beta^{\text{mul}}\) decreases computation speed but increases the electricity cost (objective of the problem) and vice versa. 

\begin{figure*}[t]
        \centering
            \subfigure[Difference in the RM at and predicted generation weather conditions for Nov \(4^{\text{th}}\) 2020]
            {
                \label{subfig:RM differences}
                \includegraphics[width=0.29\textwidth]{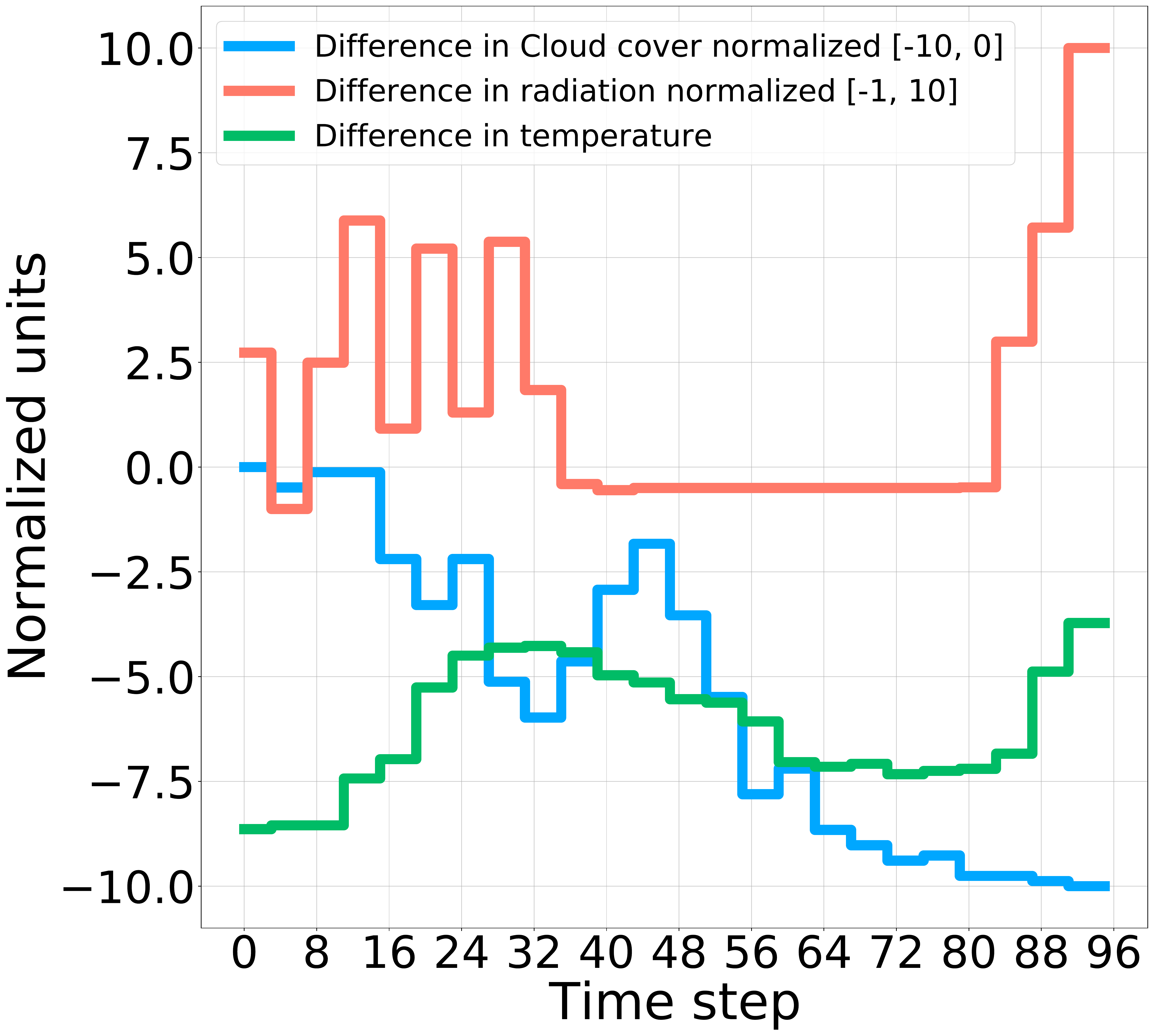} 
            } 
            \subfigure[RM vs. predicted solar generation for Nov \(4^{\text{th}}\) 2020]
            {
                \label{subfig:RM output}
                \includegraphics[width=0.29\textwidth]{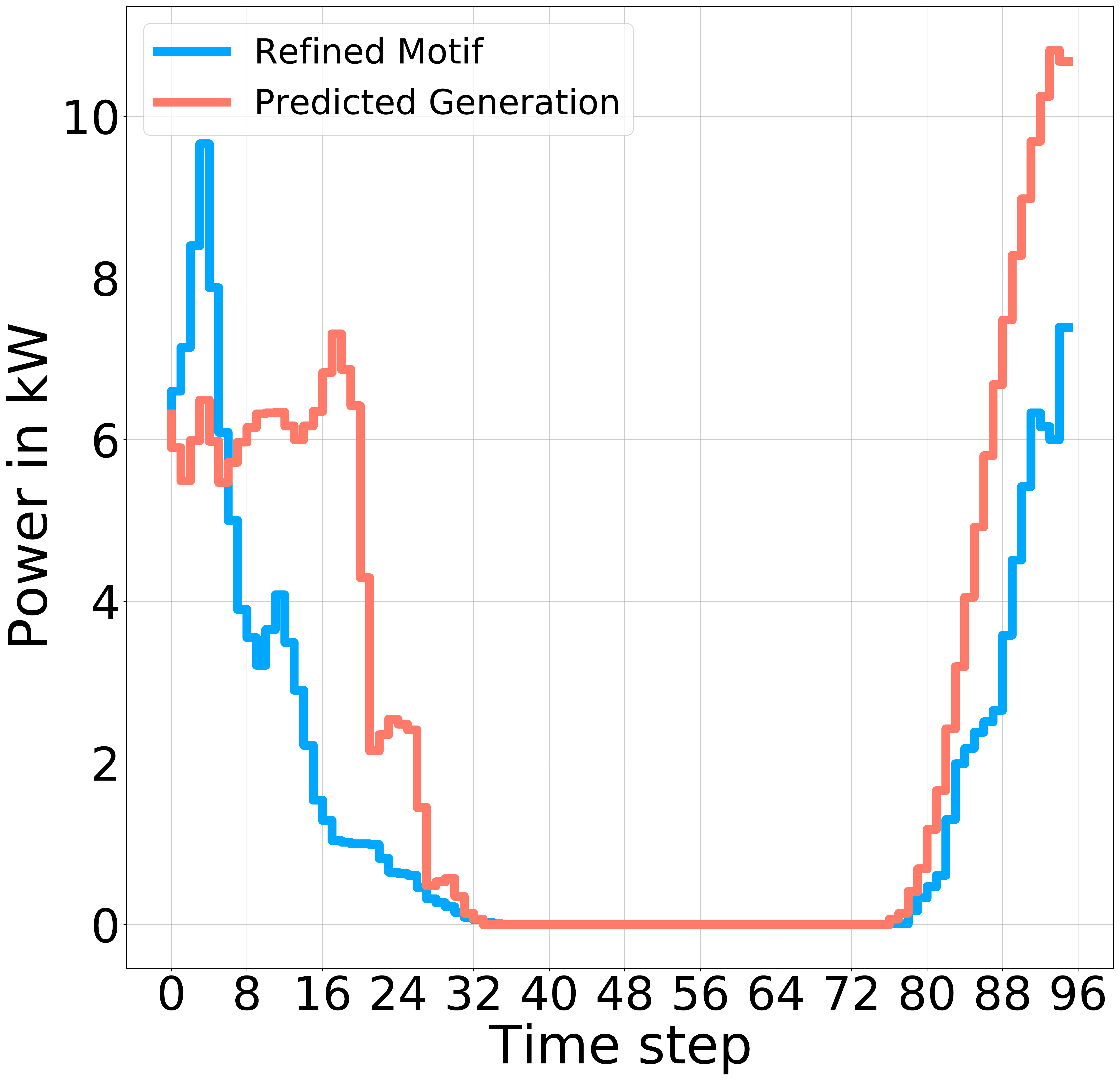} 
            } 
            \subfigure[Scheduled power, predicted net load and NEM spot price for Nov \(2^{\text{nd}}\) 2020]
            {
                \label{subfig:Optimization output}
                \includegraphics[width=0.33\textwidth]{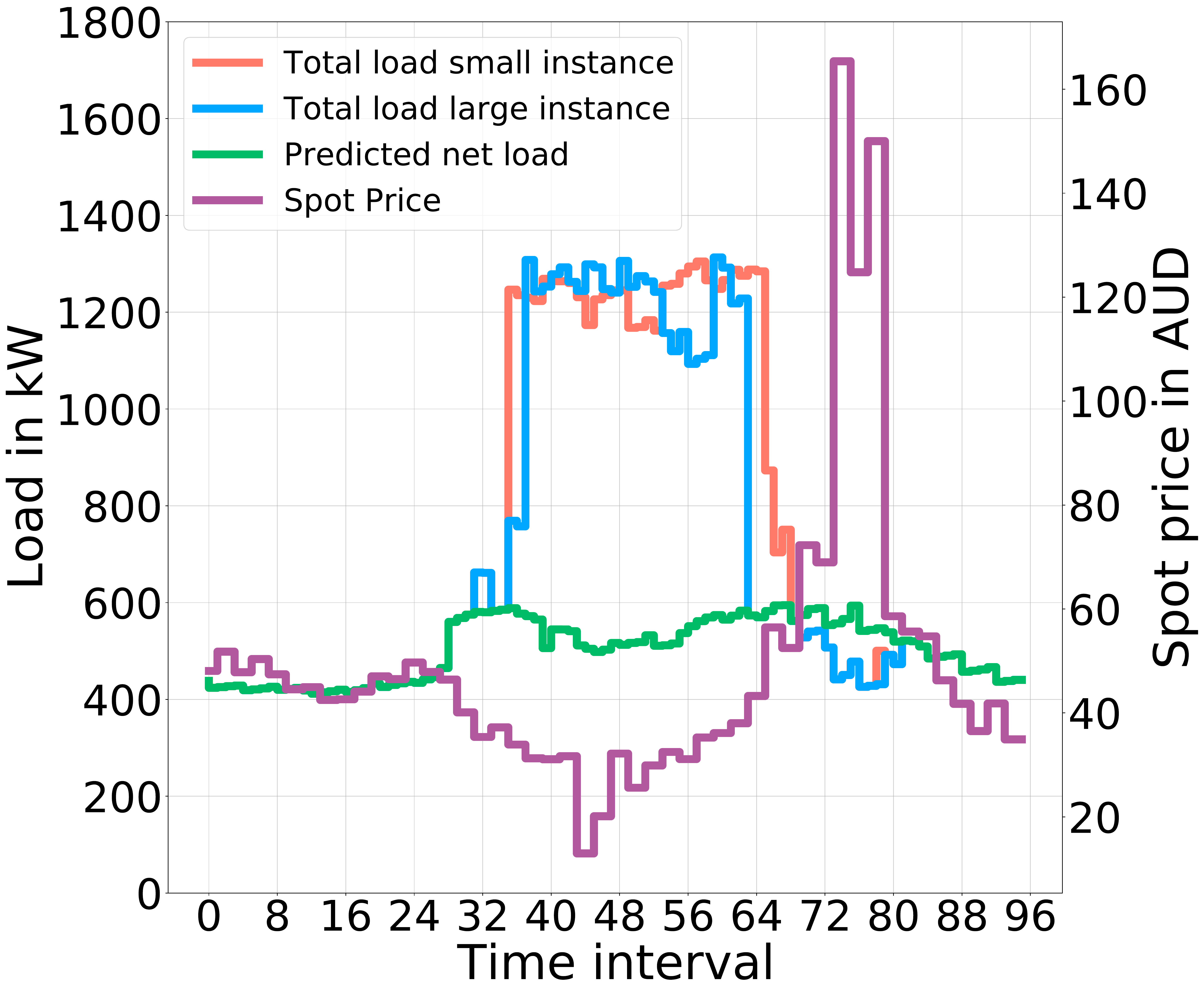}
            }
        \vspace{-0.4em}
        \caption{Prediction and optimization results}
        \label{fig:Figure ref}
\end{figure*}

\section{Results and Discussion}
\label{sec:Results and Discussion}

The simulations were run for November 2020 as per the competition stipulation with \(\Delta t = 15 \text{ min}, \
\|\mathcal{T}\|=2880\) and other instance information as specified in \ref{subsec:Instance}. For the scheduling, two \(\beta^{\text{mul}}\) values were chosen; 1.10 for small instances and 1.15 for large instances. These values were obtained via trial and error. The prediction and optimization algorithms were run in Python 3.8.8 using Gurobi\textsuperscript{\textregistered} solver. 

\begin{table}[t]
    \caption{Baseload and PV generation prediction NRMSE}
    \centering
    \begin{tabular}{ccc|ccc}
    \toprule
    \rowcolor{AdOrange} \textbf{Building \#} & \textbf{Proposed} &
    \textbf{BI} & \textbf{Solar \#} & \textbf{Proposed} & \textbf{BI} \\
    \midrule
    0 & \textbf{0.43} & 0.43 & 0 & \textbf{0.72} & 0.81 \\
    \rowcolor{AdOrangeLight} 1 & \textbf{0.31} & 0.34 & 1 & \textbf{0.77} & 0.77 \\
    2 & \textbf{0.23} & 0.24 & 2 & \textbf{0.74} & 0.79 \\
    \rowcolor{AdOrangeLight} 3 & \textbf{0.32} & 0.32 & 3 & \textbf{0.74} & 0.75 \\
    4 & \textbf{0.40} & 0.40 & 4 & \textbf{0.78} & 0.79 \\
    \rowcolor{AdOrangeLight} 5 & \textbf{0.13} & 0.14 & 5 & \textbf{0.72} & 0.75 \\
    \bottomrule
    \end{tabular}
    \label{tab:mae}
\end{table}

Since each of the twelve datasets has different scales, we decide to use the normalised root mean square error (NRMSE), which has been used to facilitate the comparison among time series with varying magnitudes \cite{das2018evaluation}. The metric is calculated as follows:
\begin{flalign}
\label{eq:NRMSE}
& \text{NRMSE} = \sqrt{\sum_t \frac{\hat{y}_t - y_t}{\|\mathcal{T}\|}} \times \frac{1}{\bar{y}}
\end{flalign}
where $\hat{y}_t$ is the model predictions, $y_t$ is the model actual data and $\bar{y}$ represents the mean of the actual data.

In order to demonstrate the superiority of our methods, we compare our proposed techniques with the best individual (BI) prediction models that are mentioned in section \ref{subsec:Related Work}. Please note that the BI models are picked from out-of-sample validation on the actual datasets using the same inputs and training length as the proposed methods. Table~\ref{tab:mae} displays the NRMSE of the actual data in November 2020 between the proposed and the BI models. It can be seen that the proposed models achieve lower errors in almost all the twelve profiles except for the demand of buildings 0 and 4. As mentioned in section \ref{subsec:baseload forecast}, these two profiles are predicted using either RF or GB due to their non-cyclic behavior. Therefore, they have the highest errors among all the demand profiles. For the PV generation time series, the proposed method also achieves better prediction accuracy with quite consistent NRMSE due to the high dependence of solar PV generation on weather data.

Fig.~\ref{subfig:RM differences} and \ref{subfig:RM output} are used to explain the intuition behind the RM method for the PV generation; the former shows the difference between RM 
and a predicted generation weather condition for November 4\textsuperscript{th}, 2020  while the latter shows the RM and the predicted generation profiles for the same day. We can see that as the difference in temperature is positive, the predicted generation is lower than the RM and vice versa. The ResNet is then trained with the changes in weather conditions to learn how to reshape the RM to predict solar generation.

Fig.~\ref{subfig:Optimization output} shows the power scheduled for the small and large instances, the predicted load and the NEM spot price profiles for Novemeber \(2^{\text{nd}}\) 2020. It can be seen that during lower price periods (intervals 24--72), most of the load is scheduled and the scheduled load is clipped to the value obtained from sub-problem 1. During high price periods (intervals 72--88), the scheduled load is lower than the predicted baseload due to battery operation. This ensures that electricity cost is minimized. 

\begin{table}[b]
    \caption{Electricity cost from different forecasting results}
    \centering
    \begin{tabular}{c|ccc}
    \toprule
    \rowcolor{AdOrange} \textbf{Forecast methods} &
    \textbf{Base case} & \textbf{BI} & \textbf{Proposed} \\
    \midrule
    \rowcolor{AdOrangeLight} \textbf{Cost (\$AUD)} & 589356 & 466652 & \textbf{357210} \\
    \bottomrule
    \end{tabular}
    \label{tab:opti_cost}
\end{table}

The organizers provided a base schedule/electricity cost for the problem based on no predictions and an \(\epsilon\)-greedy algorithm \cite{IEEECIS}. Table~\ref{tab:opti_cost} compares the electricity cost from three different forecast methods, where the base case represents the electricity cost provided by the organizers; whereas BI and the proposed cost represent the cost optimized on their respective forecasts. Our methodology reduced this cost by roughly 23\% and 40\% compared to the BI and base case, respectively. We were able to solve small instances (158400 binaries) in an average time of 447 seconds per instance and large instances (590400 binaries) in 1550 seconds per instance. It can be seen that even though the problem size grew 3.73 times, the time to solve the problem increased almost linearly and not exponentially with problem size as is the case in MIPs. It further demonstrates the tractability and scalability of our algorithm. 


\section{Conclusion}
\label{sec:Conclusion}

In this paper, we developed an optimal scheduling solution for solving a real-world electricity cost minimization problem for six buildings at Monash University in Melbourne, Australia. A two-step solution was presented in this paper, which included forecasting and an optimization step. Our solution with energy forecast reduced the total cost of electricity roughly by 40\%. We would like to highlight that we placed fifth in the competition among 50 participants and were off from the best solution only by 4\% with respect to the objective value due to their better forecast. 
For future work, the forecasting of baseload and solar generation can be improved by using 
modern machine learning techniques, such as transformers and info-GANs. 
For the optimization, a major improvement can be devising an algorithm to select optimal \(\beta^{\text{mul}}\) considering the trade-off between the tightness (complexity) of the problem and cost minimization.   
\bibliographystyle{IEEEtran}
\bibliography{reference.bib}
\end{document}